\documentclass[10pt,twocolumn,letterpaper]{article}

\usepackage{3dv}
\usepackage{times}
\usepackage{epsfig}
\usepackage{graphicx}
\usepackage{amsmath}
\usepackage{amssymb}

\usepackage{booktabs}
\usepackage{threeparttable}
\usepackage{multirow}

\usepackage[pagebackref=false,breaklinks=true,letterpaper=true,colorlinks,bookmarks=false]{hyperref}

\threedvfinalcopy 

\def\threedvPaperID{77} 

\ifthreedvfinal\fi
\begin{document}

\title{Exploring Versatile Prior for Human Motion via Motion Frequency Guidance}

\author{
Jiachen Xu\textsuperscript{\rm 1}\;
Min Wang\textsuperscript{\rm 2}\;
Jingyu Gong\textsuperscript{\rm 1}\;
Wentao Liu\textsuperscript{\rm 2}\;
Chen Qian\textsuperscript{\rm 2}\;
Yuan Xie\textsuperscript{\rm 3$*$}\;
Lizhuang Ma\textsuperscript{\rm 1,3}\thanks{Corresponding Author. Lizhuang Ma is the member of Qing Yuan Research Institute, Shanghai Jiao Tong University.}\\
\\
\textsuperscript{\rm 1} Shanghai Jiao Tong University\;
\textsuperscript{\rm 2} SenseTime Research\;
\textsuperscript{\rm 3} East China Normal University\\
{\tt\small xujiachen@sjtu.edu.cn\;\quad  wangmin@sensetime.com\;\quad gongjingyu@sjtu.edu.cn\;\quad}\\
{\tt\small \{liuwentao,qianchen\}@sensetime.com\;\quad yxie@cs.ecnu.edu.cn\;\quad ma-lz@cs.sjtu.edu.cn}
}

\maketitle
\thispagestyle{empty}

\begin{abstract}
   Prior plays an important role in providing the plausible constraint on human motion. Previous works design motion priors following a variety of paradigms under different circumstances, leading to the lack of versatility. In this paper, we first summarize the indispensable properties of the motion prior, and accordingly, design a framework to learn the versatile motion prior, which models the inherent probability distribution of human motions. Specifically, for efficient prior representation learning, we propose a global orientation normalization to remove redundant environment information in the original motion data space. Also, a two-level, sequence-based and segment-based, frequency guidance is introduced into the encoding stage. Then, we adopt a denoising training scheme to disentangle the environment information from input motion data in a learnable way, so as to generate consistent and distinguishable representation. Embedding our motion prior into prevailing backbones on three different tasks, we conduct extensive experiments, and both quantitative and qualitative results demonstrate the versatility and effectiveness of our motion prior. Our model and code are available at \href{https://github.com/JchenXu/human-motion-prior}{https://github.com/JchenXu/human-motion-prior}.
\end{abstract}

\section{Introduction}
\label{sec:intro}
Human motion modeling plays an important role in many applications such as video games and computer animation. 
Many tasks study the human motion under different circumstances~\cite{zhang2019predicting,kocabas2020vibe,mao2020history,zhang2021we,ling2020character}. Most of them ask for the physically plausible human motion, which means that any independent pose of the motion is plausible, as well as the transition between poses is reasonable.

Several works study pose priors by exploring the constraint on human poses~\cite{bogo2016keep,kanazawa2018end,pavlakos2019expressive,zanfir2020weakly}. 
However, a plausible motion asks for both the continuity between poses and the feasibility of the independent pose. 
Hence, recent works try to design the prior for human motion. Kocabas~\etal~\cite{kocabas2020vibe} design an adversarial prior to discriminate between generated and real human motions so as to keep the predicted motion plausible. In addition, Holden~\etal~\cite{holden2016deep} learn a motion manifold, where each motion data is embedded into a low-dimensional representation.
However, they design the motion prior only for specific tasks. We argue that a pre-trained and task-agnostic human motion prior is essential for those motion generation tasks because of the insufficient paired 3D motion data. Therefore, we summarize the indispensable properties of the motion prior, and design a versatile motion prior according to these properties.

\begin{figure}
    \centering
    \includegraphics[width=\linewidth]{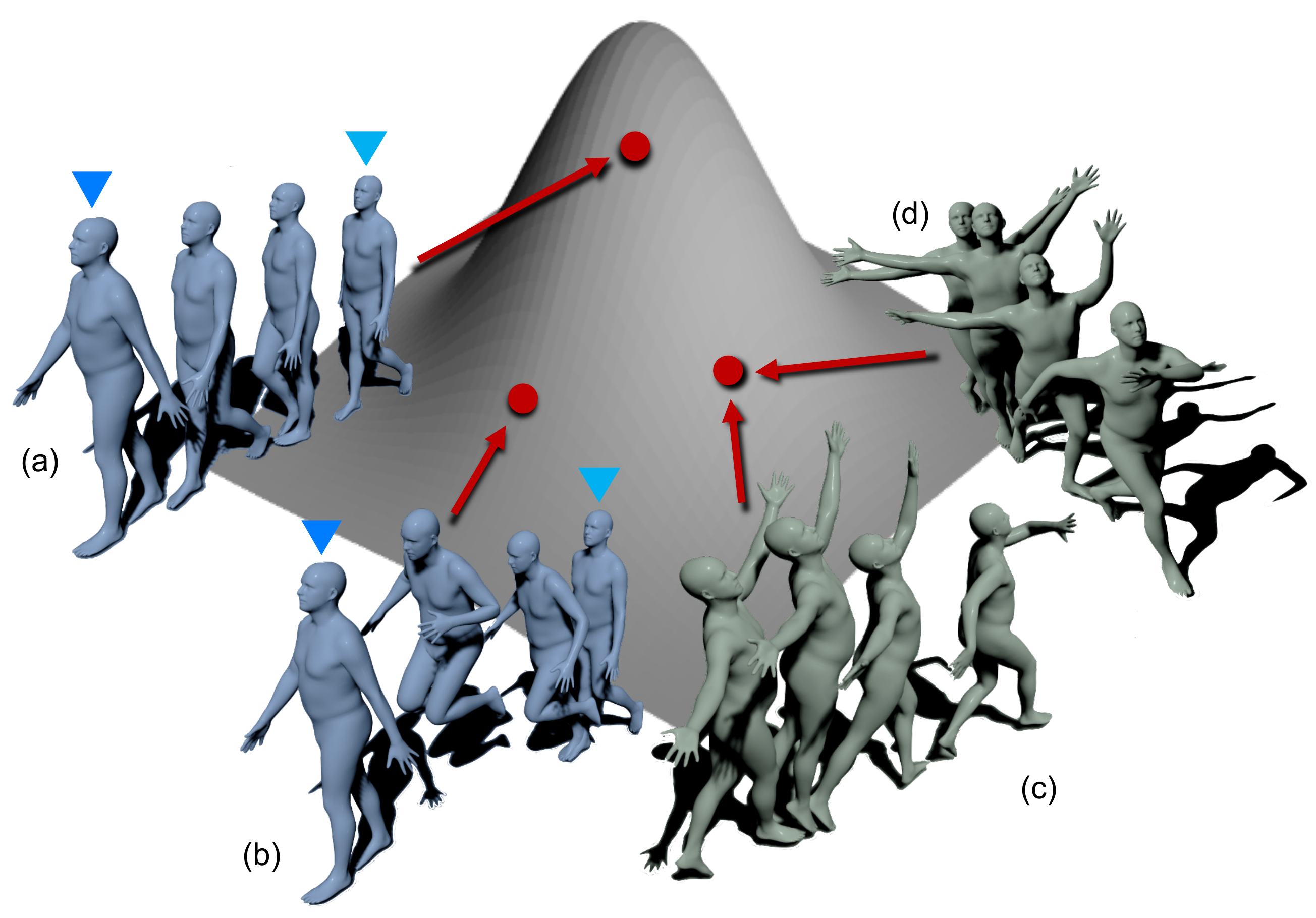}
    \caption{Illustration of our learnt motion prior representation space. Given an undersampled observation (\emph{i.e.,} start and end position of (a) and (b) pointed by blue arrow), there may be multiple solutions but with different probabilities, which help to solve the ambiguity. Meanwhile, same motion sequence but with different global orientations, (c) and (d), have the consistent representation in our motion prior space.}
    \vspace{-3mm}
    \label{fig:fig1}
\end{figure}

First, \textit{a tractable and continuous distribution} over the latent representation is required to model the inherent probability distribution of human motions. It is important for solving ambiguity in ill-posed tasks, such as motion infilling and prediction. Because some common motions have a high probability of occurrence, while the probability of rare or impossible motions is lower. 
For example, in Fig.~\ref{fig:fig1}, the full motion (a) and (b) have the same undersampled observation (\textit{i.e.,} start and end position pointed by the blue arrow), leading to the ambiguity in the ill-posed motion infilling. If the prior models the probability distribution of the human motion data, the ambiguity can be solved by offering a more probable solution which is more likely to conform to human behavior. 
Therefore, we construct a motion prior based on the probabilistic model, variational auto-encoder (VAE)~\cite{kingma2013auto}, to model the inherent motion distribution.


Second, \textit{a complete and efficient space} is significant for constructing a versatile prior, since the prior needs to generalize to various tasks and datasets without fine-tuning.
Specifically, the complete and efficient space means that we can accurately reconstruct any plausible motion from a low-dimensional representation. A large-scale dataset with a variety of long-term motions is usually adopted for the completeness. However, it leads to a complex data space that is hard to be represented in low dimension. So, we propose to learn an efficient representation space from two aspects: reducing the complexity of the data space to be modeled and encoding each motion data efficiently.

For the reduction of complexity of the data space, Luo~\etal~\cite{luo20203d} resort to shorter-term motions to be modeled. However, more context information provided by long-term motion is beneficial for down-stream tasks. By contrast, we introduce a global orientation normalization, which normalizes the global orientation of each motion around yaw axis while retaining the relative orientation transition between frames. Since the direction in which a person moves is related to the environment, the global orientations of motion in the dataset are biased on the environment information. So it is non-trivial to remove the redundant environment information in the data space as well as reducing the complexity.

For the efficient motion encoding, motion segments with slower changes between frames should be efficiently compressed, while motion segments with higher degree of variation deserve more attention. For instance, in a motion sequence, the dancer may stand still for a while before dancing. Compared with standing segments, dancing segments carry more information and details, and deserve to be well retained~\cite{xiao2020invertible}. Thus, we introduce a two-level motion frequency guidance
to efficiently encode the motion into the low-dimensional prior representation. One level is sequence-based and the other is segment-based. The sequence-based frequency guidance captures the difference between frequency patterns of motions and provides category cues from the frequency~\cite{hu2019skeleton}. The segment-based frequency guidance exploits the frequency difference between segments within a motion to adaptively compress segments with different amount of high-frequency information. 

Third, \textit{a consistent and distinguishable representation} should be learnt in the motion prior.
As aforementioned, the same motion, consisting of same poses and relative transitions, may have different global orientations as the environment changes. For example, in Fig. \ref{fig:fig1}, (c) and (d) correspond to the same motion with different orientations, but they are supposed to have the consistent representation. A straightforward solution is to take our orientation normalized motions as both input and output of our motion prior, so as to explicitly remove the global orientation for each input motion during training and inference phase. Yet, this solution may cause the prior fail to capture underlying distinguishable features for the motion.
Thus, we introduce a denoising training scheme to disentangle the global orientation (environment information) from the human motion data in a learnable way, so as to learn a consistent representation while keeping the representation distinguishable~\cite{bengio2013representation,im2017denoising}. 

To demonstrate the versatility and effectiveness, we integrate our pre-trained motion prior into different backbones without fine-tuning for different tasks, such as human motion reconstruction, motion prediction and action recognition. Then, we conduct experiments on 3DPW~\cite{von2018recovering}, Human3.6M~\cite{ionescu2013human3} and BABEL~\cite{BABEL_CVPR_2021} to evaluate the performance on different tasks. Results show that our motion prior improves the baseline and achieves the state-of-the-art performance on all three benchmarks.

In summary, our contributions in this paper are:
(1) We first summarize three indispensable properties for the motion prior to achieve versatility, and accordingly,
(2) We introduce the global orientation normalization and a two-level motion frequency guidance to learn the versatile motion prior with a denoising training scheme.
(3) We integrate the proposed prior into prevailing backbones and achieve the state-of-the-art performance on different benchmarks, which demonstrates the versatility of our motion prior.

\section{Related Work}
\label{sec:related_work}

\textbf{Human pose and motion prior.} Constructing a kind of prior is commonly used in pose~\cite{bogo2016keep,kanazawa2018end,pavlakos2019expressive,zanfir2020weakly} and motion~\cite{urtasun20063d,holden2016deep,kocabas2020vibe,luo20203d} modeling.
Pavlakos~\etal~\cite{pavlakos2019expressive} utilize VAE~\cite{kingma2013auto} to build a non-linear manifold as the human pose prior and provide plausibility constraint. Zanfir~\etal~\cite{zanfir2020weakly} construct a wrapped prior space with normalizing flow. Compared with pose priors, motion priors have constraints on both independent poses and transition between poses. Kocabas~\etal~\cite{kocabas2020vibe} train an adversarial discriminator as motion prior to discriminate between generated motions and real human motions. 
Holden~\etal~\cite{holden2016deep} employ the autoencoder to encode all plausible motion into a compact manifold, where each latent code can represent a plausible human motion. Luo~\etal~\cite{luo20203d} try to compress a large-scale dataset, AMASS~\cite{mahmood2019amass}, with VAE into a representation space. In this paper, we first analyze the properties of a versatile motion prior and the characteristics of human motion itself. Then, we accordingly propose the global orientation normalization and a two-level motion frequency guidance.

\textbf{Frequency in motion modeling.} Previous works convert the motion into frequency domain and take the frequency coefficients as input to combine both spatial and temporal information~\cite{mao2019learning,cai2020learning}. Mao~\etal~\cite{mao2020history} represent historical sub-sequences of each motion as frequency coefficients, and aggregate them with attention mechanism to predict the future motion. Zhang~\etal~\cite{zhang2021we} encode the motion into different DCT spaces to decompose the motion into several frequency bands. By contrast, we exploit the characteristic of frequency that it represents the amount of information, to adaptively compress the human motion data.

\textbf{Human motion reconstruction.}
Reconstructing the 3D human pose or motion has attracted significant interest~\cite{bogo2016keep,kanazawa2018end,kolotouros2019learning,song2020human,kanazawa2019learning,kocabas2020vibe,zanfir2020weakly}. 
Compared with poses, reconstructing human motion has more demanding requirements for shape consistency and smoothness. Kanazawa~\etal~\cite{kanazawa2019learning} present a temporal encoder to learn 3D human dynamics from video and generate smooth motion. Kocabas~\etal~\cite{kocabas2020vibe} design a GRU-based network with the adversarial prior to guide the motion inference.
Choi~\etal~\cite{choi2021beyond} explicitly exploit the past and future frames to achieve smoother and better results. Meanwhile, \cite{shimada2020physcap,yuan2021simpoe,rempe2020contact,PhysAwareTOG2021} try to improve the physical fidelity of generated motion through reinforcement learning and other physical constraints.

\textbf{Motion Prediction.} Motion prediction from past pose sequence is studied from two aspects: deterministic~\cite{aksan2019structured,cui2020learning,mao2020history,zhang2019predicting} and stochastic~\cite{zhang2020perpetual,yan2018mt,yuan2020dlow,aliakbarian2020stochastic}. Instead of using a sequence of past poses, Chao~\etal~\cite{chao2017forecasting} forecast human dynamics from static images and Yuan~\etal~\cite{yuan2019ego} predict future motions from egocentric videos. Zhang~\etal~\cite{zhang2019predicting} utilize the SMPL model to represent the human body and predict future motions with pose and shape from videos. 

\textbf{Action Recognition.} To understand human motion, skeleton-based action recognition attracts much attention~\cite{shi2019two,cheng2020skeleton,liu2020disentangling,cheng2020eccv}. Most of them carefully design and train a GraphConv network in a supervised way on the widely-used NTU-RGBD~\cite{shahroudy2016ntu}, which has two major problems: the discontinuity of action and the lack of modeling the long-tailed distribution. Recently, the BABEL~\cite{BABEL_CVPR_2021} dataset is proposed to tackle these two problems, which is closer to the real life. 

By contrast, we take the learnt prior representation, referring to a plausible motion, as the intermedia to generate the final outputs, which benefits from the context and probability information encoded in the prior.

\section{Motion Prior}
\subsection{Human Motion Representation}
\label{sec:motion-represent}
We use the parametric human model, SMPL~\cite{loper2015smpl}, to represent each human pose in the motion. SMPL model, which can be regarded as a differential function $\mathcal{M}(\cdot)$, parameterizes the human body pose and shape through $\theta\in\mathbb{R}^{72}$ and $\beta\in\mathbb{R}^{10}$, respectively. Pose parameter $\theta$ consists of
global orientation $\theta^g$ and local body pose $\theta^l$ determined by relative rotation of 23 joints
in axis-angle format. Given $\theta$ and $\beta$, $\mathcal{M}(\theta, \beta)$ outputs a triangulated mesh with $N=6,980$ vertices. Then, we denote the motion sequence with $K$ frames as $\mathbf{X}=\{\Theta_i\}^{K}_{i=1}$, where $\Theta_i=(\theta_i, \beta_i)$ represents the human model for $i$-th frame. 

\textbf{Global Orientation Normalization.}
\label{subsec:rotation_norm}
As aforementioned, the global orientation around yaw axis of each motion in the dataset is related to the environment, while we argue that the motion prior should focus on the human motion itself. Hence, to remove the redundant environment information in the motion data and reduce the complexity of data space, we propose the global orientation normalization. 

As shown in Fig. \ref{fig:motion_prior}, we normalize the orientation of entire motion around yaw axis according to the first frame while remaining the internal relative orientation transition, so as to make all input sequences start in the same forward direction. Specifically, given an input motion sequence $\mathbf{X}$, we first normalize the first frame by clipping the yaw rotation and generate normalized orientation $\hat{\theta_1^g}$. Then we generate the correction rotation $\mathcal{R}_{cor}$ between the original orientation $\theta_1^g$ and $\hat{\theta_1^g}$ of the first frame, and normalize the global orientation of the motion sequence as follows: 
\begin{align}
    \mathcal{R}_{cor} = {\mathbf{\hat{\theta_1^g}}} & \cdot {\mathbf{{\theta_1^g}}}^{-1}= {\mathbf{\hat{\theta_1^g}}} \cdot {\mathbf{{\theta_1^g}}}^{T}, \\
    {\mathbf{\hat{\theta_i^g}}} = &\mathcal{R}_{cor} \cdot {\mathbf{\theta_i^g}},
\end{align}
where $\mathbf{\theta^g}$ is the rotation matrix format of ${\theta^g}$.

\subsection{Frequency Guiding Prior Framework}
\label{subsec:motion_prior_framework}
To construct our motion prior, we exploit the variational auto-encoder (VAE)~\cite{kingma2013auto} and learn a $256$-dimensional latent representation space. We first introduce the input data, then the encoder where we perform the two-level frequency guidance. Finally, we will introduce the decoder.

\textbf{Input data.} 
Given a set of pose and body parameters, $\theta$ and $\beta$, human motion can be expressed through SMPL model. However, the relative rotation and global orientation is less intuitive and straightforward. Therefore, we also take the joint sequence $\mathcal{J}\in\mathbb{R}^{K\times J\times3}$ of each motion as input, where $J$ is the number of body joints. Also, we explicitly calculate velocity $\mathcal{J}^{vel}$ and acceleration $\mathcal{J}^{acc}$ for each joint to better reveal the dynamic features.

Following~\cite{pavlakos2019expressive}, we also ignore the variance of shape information and take the same shape for each motion. Therefore, the motion input used to construct our motion prior is denoted as $\Phi=\{(\hat{\theta^g_i}, \theta^l_i, \beta, \mathcal{J}_i, \mathcal{J}_i^{vel}, \mathcal{J}_i^{acc})\}^K_{i=1}$.

\begin{figure}
    \centering
    \includegraphics[width=\linewidth]{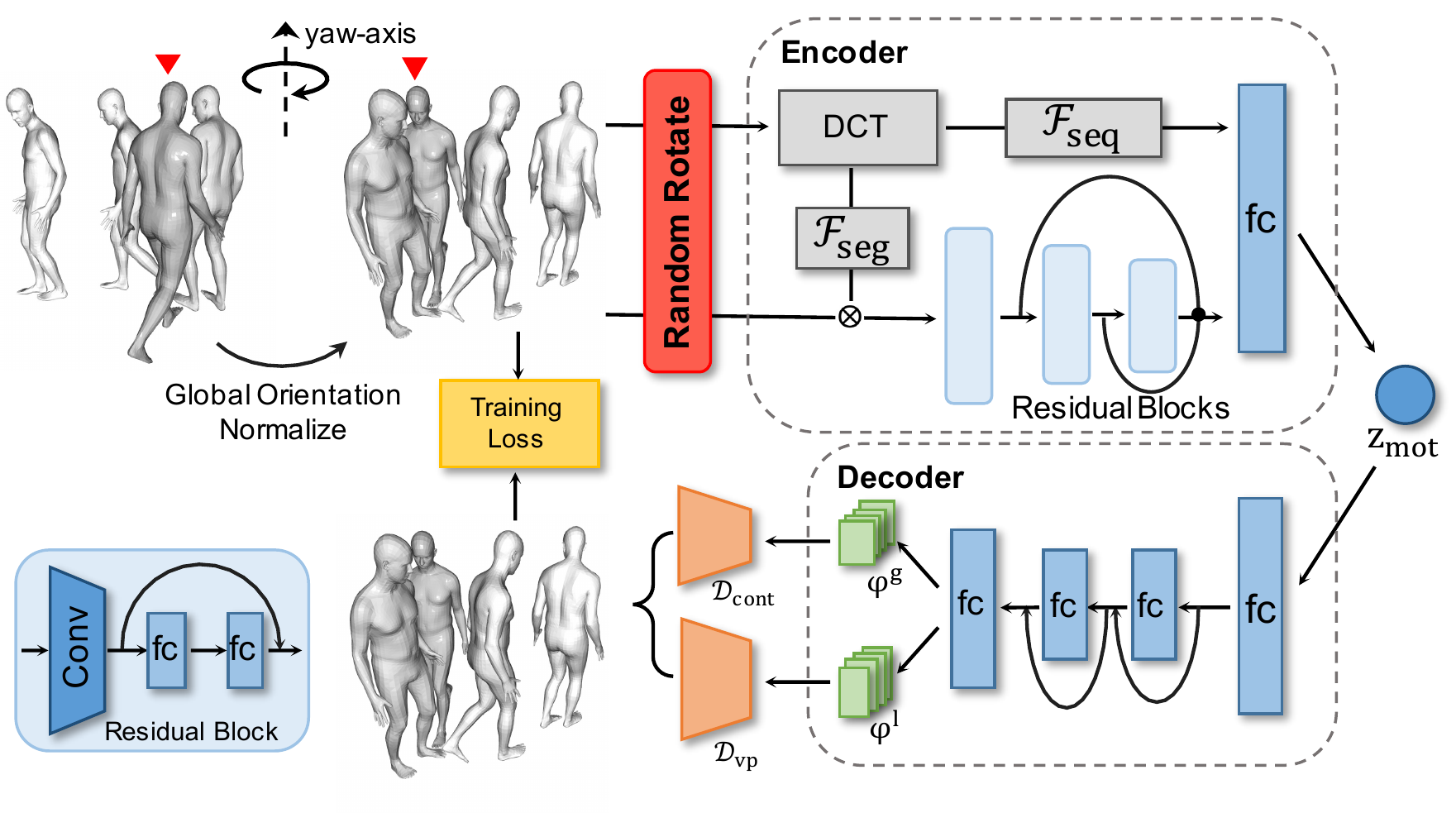}
    \caption{Pipeline and framework. We visualize the motion after global orientation normalization (the red arrow points at the beginning frame) and illustrate the framework of our motion prior.}
    \label{fig:motion_prior}
\end{figure}

\textbf{Encoder.} 
RNN-based networks, which mainly focus on temporal correlations, usually fail to capture the spatial-temporal dynamics in human motion~\cite{li2018convolutional}. Hence, as shown in Fig.~\ref{fig:motion_prior}, we construct a convolutional encoder, which takes $\Phi$ as input and consists of three residual blocks to extract both the fine-grained information and global context. 

To introduce the sequence-based and segment-based frequency guidance for efficient representation learning, we take the joint sequence $\mathcal{J}\in\mathbb{R}^{K\times J\times3}$ and further divide $\mathcal{J}$ into $S$ segments of length $n$, \textit{i.e.,} $\dot{\mathcal{J}}\in\mathbb{R}^{S\times n\times J\times3}$. Then, for each motion, we make use of the discrete cosine transform (DCT) to extract the sequence-based frequency components $\mathcal{F}_{seq}\in\mathbb{R}^{C_m\times J\times3}$ from $\mathcal{J}$ and the segment-based frequency $\mathcal{F}_{seg}\in\mathbb{R}^{S \times C_s \times J\times3}$ from $\dot{\mathcal{J}}$, where $C_m$ and $C_s$ represent the number of kept frequency components. 

Then, to perform the segment-based frequency guidance, we extract the segment attention value $\alpha_{seg}\in\mathbb{R}^{S}$ from $\mathcal{F}_{seg}$ and re-weight the segment features $f_{seg}$ extracted by the residual block for each segment, so as to adaptively compress the information according to the frequency:

\begin{equation}
    f_{seg}^{'} = f_{seg}\cdot\alpha_{seg} = f_{seg}\cdot\sigma(\phi(\mathcal{F}_{seg})),
\end{equation}
where $f_{seg}^{'}, f_{seg}\in\mathbb{R}^{S\times n\times c}$, $n$ and $c$ are the length of each segment and the channel number of the feature. $\sigma(\cdot)$ and $\phi(\cdot)$ are the softmax function and multi-layer perceptron and are used to predict the segment attention value $\alpha_{seg}$. Besides, this compressing process is conducted in the first layer for efficient compression for the entire motion.

Furthermore, we combine features from different scales as the motion dynamic feature to keep both the global context information and the fine-grained local pose information. Also, we exploit the sequence-based frequency $\mathcal{F}_{seq}$ and introduce the global motion category information from $\mathcal{F}_{seq}$ into the dynamic feature. Finally, we encode both category information and dynamic feature into the latent representation $z_{mot}\in\mathbb{R}^{256}$ with re-parameterization trick~\cite{kingma2013auto}. 

\textbf{Decoder.} Different from the encoder, an overly complex decoder may hurt the test log-likelihood and cause the overfitting~\cite{cremer2018inference,vahdat2020NVAE}. Therefore, as shown in Fig. \ref{fig:motion_prior}, our decoder consists of two residual blocks containing one fully connected layer each. The final layer outputs the reconstructed normalized global orientation $\varphi^g$ represented by 6D continuous rotation feature~\cite{zhou2019continuity} and a latent local pose representation $\varphi^l \in \mathbb{R}^{32}$, in VPoser latent space~\cite{pavlakos2019expressive}, which is a reasonable sub-manifold for human pose. Furthermore, $\mathcal{D}_{cont}$ converts $\varphi^g$ to the axis-angle format and the decoder $\mathcal{D}_{vp}$ of VPoser~\cite{pavlakos2019expressive} with pre-trained and fixed weights decodes $\varphi^l$ into predicted local body pose $\Bar{\theta^l}$ in axis-angle format.

\subsection{Denoising Training Scheme}
To learn a consistent and distinguishable representation for the same motion, we design a denoising training scheme. 
Given a motion sample $\Phi$ after orientation normalization, we randomly apply a rotation around yaw axis to it, which can be regarded as an inverse process of normalization in Sec.~\ref{subsec:rotation_norm} with random degree, and generate a corrupted sample $\tilde{\Phi}$. Then, our prior is trained to reconstruct normalized motion $\{\hat{\theta^g},\theta^l\}$ from $\tilde{\Phi}$ instead of $\Phi$ as follows:
\begin{align}
    \mathcal{L} = \lambda_{rec}\mathcal{L}_{rec} &+ \lambda_{kl}\mathcal{L}_{kl} + \lambda_{vposer}\mathcal{L}_{vposer}\label{eq:total-loss},\\
    \mathcal{L}_{rec} = \mathcal{M}(\hat{\theta^g}||\theta^l, \beta)&-\mathcal{M}(\mathcal{D}_{cont}(\tilde{\varphi}^g)||\mathcal{D}_{vp}(\tilde{\varphi}^l), \beta)\label{eq:rec},\\
    &(\tilde{\varphi}^g,\tilde{\varphi}^l) = \mathcal{F}(\tilde{\Phi})\label{eq:enc},\\
    \mathcal{L}_{KL} = &KL(q(z_{mot}|\tilde{\Phi})||\mathcal{N}(0, I))\label{eq:kl},\\
    &\mathcal{L}_{vposer} = |\varphi^l| ^ 2\label{eq:vpose},
\end{align}
where Eq.~(\ref{eq:rec}) is the reconstruction loss for the mesh vertices of SMPL model $\mathcal{M}$ under the same shape parameter $\beta$.
Eq.~(\ref{eq:enc}) denotes the reconstruction from $\tilde{\Phi}$ through our proposed motion prior framework $\mathcal{F}(\cdot)$.
The Kullback-Leibler divergence given by Eq.~(\ref{eq:kl}) encourages the approximate posterior to be close to the normal distribution. Eq.~(\ref{eq:vpose}) constrains the human body pose to the natural range. 
\section{Versatility of Motion Prior}
\begin{figure}
    \centering
    \includegraphics[width=\linewidth]{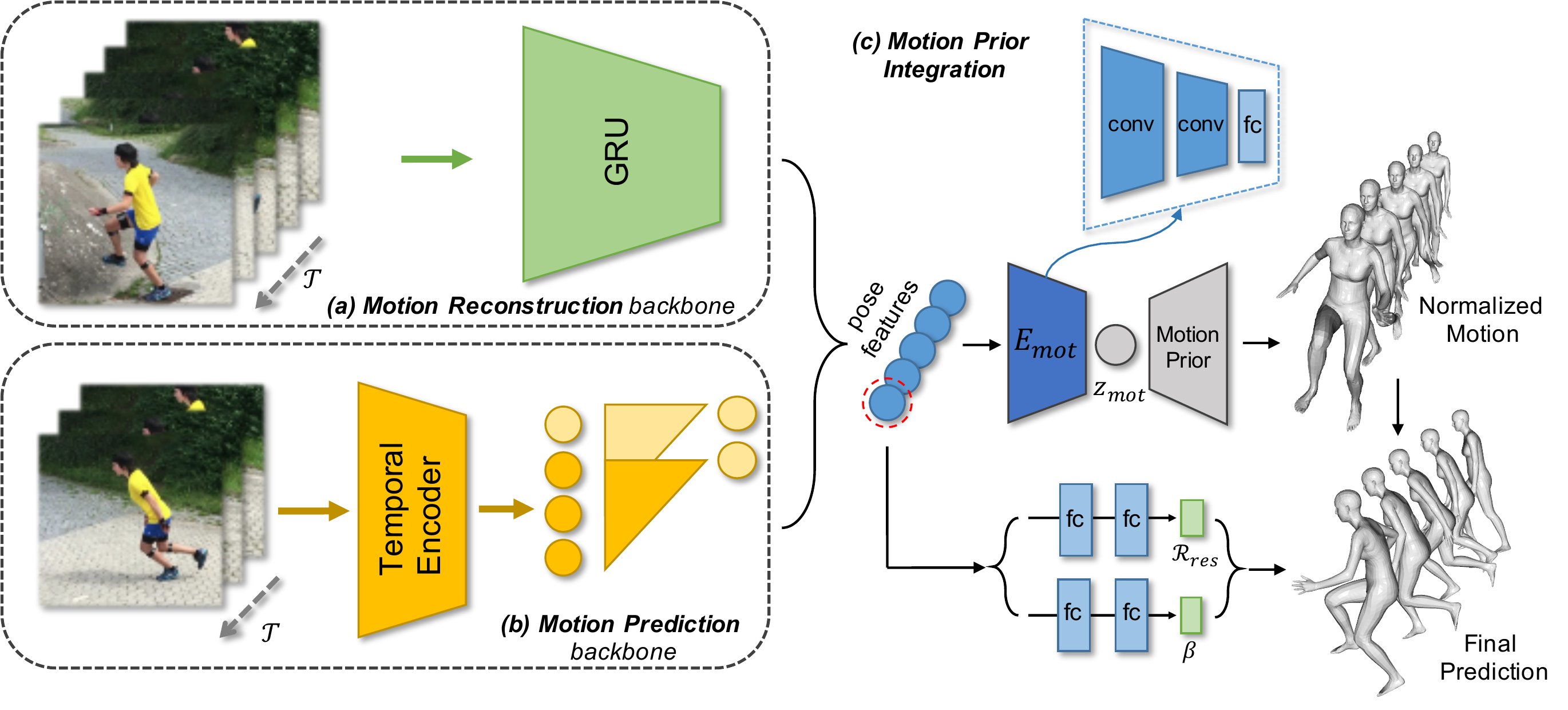}
    \caption{Integration of our proposed motion prior. We integrate the pre-trained motion prior with fixed weights into (a) VIBE in a motion reconstruction task and (b) PHD in a motion prediction task, which are simply illustrated.}
    \label{fig:reconstruct_frame}
\end{figure}
To demonstrate the versatility and effectiveness of our proposed motion prior, in this section, we integrate our motion prior into several prevailing backbones in different human motion modeling tasks.
\subsection{Human Motion Reconstruction}
\label{subsec:3d-reconstruct}
\textbf{Problem definition.} Given a video sequence $\{I_t\}_{t=1}^T$, we reconstruct the 3D human pose and shape $\{\Theta_t\}_{t=1}^T$ (defined the same as Sec. \ref{sec:motion-represent}) from each frame. It is noteworthy that the reconstructed shape parameter $\beta_t$ should be consistent across the whole sequence for a person.

\textbf{Architecture.} We utilize the VIBE~\cite{kocabas2020vibe} as our backbone and embed our motion prior into it as shown in Fig. \ref{fig:reconstruct_frame}. VIBE extracts temporal feature for each frame through Gated Recurrent Units (GRU). Then, for each frame, they produce pose $\theta$, shape $\beta$, and scale and translation $[s; t]$ of camera using a shared regressor~\cite{kanazawa2018end} from each temporal feature. 

However, we predict the pose for each frame in the sequence at once from our motion prior, instead of predicting frame-wisely through the regressor. As illustrated in Fig. \ref{fig:reconstruct_frame}, we construct a motion encoder $\mathit{E}_{mot}$, consisting of two convolutional layers and a fully-connected layer, to predict the motion representation $z_{mot} \in \mathbb{R}^{256}$. Then, the pre-trained motion prior decodes $z_{mot}$ into the motion with $K$ frames. However, the length $T$ of input video is required to be less than $K$ and we use the first $T$ poses $ \{(\Bar{\theta_i^g}, \Bar{\theta_i^l})\}_{i=1}^T$ as the output. Compared with producing poses for consecutive frames one by one, our motion prior provides more context information between poses for accurate prediction. Also, we discard the motion discriminator in VIBE, which acts as a prior to generate plausible motion but fails to solve ambiguity. By contrast, our motion prior generates more probable motion to solve ambiguity while keeping plausibility.

In addition, due to the global orientation normalization (see Sec. \ref{subsec:rotation_norm}), the predicted global orientation sequence $\{\Bar{\theta_t^g}\}_{t=1}^T$ has been normalized according to the first frame. Therefore, we construct another branch to predict the residual rotation $\mathcal{R}_{res}$ around yaw axis for the first frame and rectify the $\{\Bar{\theta_t^g}\}_{t=1}^T$ by $\Bar{\theta_t^g} = \mathcal{R}_{res}\cdot\Bar{\theta_t^g}.$

Furthermore, we also introduce a branch to directly predict the $\beta$ from the first frame and use the predicted $\beta$ for all rest frames to ensure the shape consistency across a video, where the regressor in VIBE may fail. However, given the 2D keypoints supervision, the camera model for each frame is still needed and we keep regressing $[s; t]$ based on predicted shape and pose from each temporal feature.

\subsection{Motion Prediction}
\label{subsec:motion_predict}
\textbf{Problem definition.} In this task, we aim to predict the future 3D human motion $\{\Theta_{T+1}, \Theta_{T+2}, ..., \Theta_{T+N}\}$ conditioning on a past 2D video sequence $\{I_1, I_2, ..., I_T\}$. 

\textbf{Architecture.} An auto-regressive framework PHD~\cite{zhang2019predicting} is taken as our backbone. 
It conducts auto-regressive prediction in the feature space, where each feature is regularized by an adversarial pose prior~\cite{kanazawa2018end}.
Given a video, PHD first extracts the temporal feature $\{f_t\}_{t=1}^{T}$ for each input frame, and then predicts $\{f_t\}_{t=T+1}^{T+N}$ in an auto-regressive manner, and accordingly generate poses for future motion. 

However, the auto-regressive manner is prone to cause compounding errors and fails to capture the long-range context information of whole sequence, which leads to unnatural and inaccurate generated motion. Therefore, we take the architecture introduced in Sec. \ref{subsec:3d-reconstruct} to impose the regularization over the whole sequence of pose features, and refine errors and unnaturalness. As shown in Fig. \ref{fig:reconstruct_frame}, we take both $\{f_t\}_{t=1}^{T}$ and the auto-regressed features $\{f_t\}_{t=T+1}^{T+N}$ as input and generate $z_{mot}$ by the motion encoder $\mathit{E}_{mot}$, which embeds the long-range context information. Then, the pre-trained prior decoder generates the human motion from $z_{mot}$, which is not only constrained to the plausible space but also refined with the context information.

\begin{table}[]
    \small
    \centering
    \begin{threeparttable}
    \resizebox{\linewidth}{!}{
    \begin{tabular}{lcccc}
        \toprule
        Method & MPJPE $\downarrow$ & PA-MPJPE $\downarrow$ & MPVPE $\downarrow$ & Accel Error $\downarrow$ \cr
        \midrule
        MEVA~\cite{luo20203d} & - & 43.9 & - & - \\
        \midrule
        Vanilla VAE & 52.82 & 17.88 & 63.06 & 5.22\\
        ~~~~~~~ $+ \mathcal{F}_{seq}$ & 34.01 & {\bf 17.18} & 41.02 & 5.24 \\
        ~~~~~~~ $+ \mathcal{F}_{seg}$ & 26.59 & 17.66 & 33.25 & 5.17 \\
        \midrule
        Ours & {\bf 26.01} & 17.44 & {\bf 32.70} & {\bf 5.07} \\
        \bottomrule
    \end{tabular}}
    \end{threeparttable}
    \vspace{1pt}
    \caption{VAE reconstruction error on 3DPW. ``$+\mathcal{F}_{seq}$" and ``$+\mathcal{F}_{seg}$" stands for the guidance of sequence-based frequency and segment-based frequency. ``Accel Error" stands for the acceleration error and ``$\downarrow$" denotes that the lower the result, the better.}
    \label{tab:prior_ablation}
\end{table}

\begin{table*}
  \small
  \centering
  \begin{threeparttable}
  \resizebox{0.98\linewidth}{!}{
    \begin{tabular}{clccccccc}
    \toprule
    &\multirow{2}{*}{Method}&
    \multicolumn{4}{c}{{3DPW}}&\multicolumn{3}{c}{{Human3.6M}}\cr
    \cmidrule(lr){3-6} \cmidrule(lr){7-9}
    & & MPJPE $\downarrow$ & PA-MPJPE $\downarrow$ & MPVPE $\downarrow$ & Accel Error $\downarrow$ & MPJPE $\downarrow$ & PA-MPJPE $\downarrow$ & Accel Error $\downarrow$\cr
    \cmidrule(lr){1-9}
    \multirow{3}{*}{\rotatebox{90}{\begin{tabular}{c}Frame \\ based\end{tabular}}} & SPIN ~\cite{kolotouros2019learning} & 96.9 & 59.2 & 116.4 & 29.8 & - & {41.1} & - \\
    & I2L-MeshNet~\cite{Moon_2020_ECCV_I2L-MeshNet} & 93.2 & 58.6 & 110.1 & 30.9 & {\bf 55.7} & 41.7 & - \\
    & Pose2Mesh~\cite{choi2020pose2mesh} & 88.9 & 58.3 & 106.3 & - & 64.9 & 46.3 & - \\
    \cmidrule(lr){1-9}
    \multirow{5}{*}{\rotatebox{90}{\begin{tabular}{c}Video \\ based\end{tabular}}} & HMMR~\cite{kanazawa2019learning} & 116.5 & 72.6 & 139.3 & 15.2 & - & 56.9 & - \\
    & Sun~\etal~\cite{sun2019human} & - & 69.5 & - & - & 59.1 & 42.4 & - \\
    & VIBE~\cite{kocabas2020vibe} & 93.9 & 55.9 & 112.6 & 27.0 & 65.6 & 41.4 & 27.3 \\
    \cmidrule(lr){2-9}
    & Ours\ddag & {85.1} & {\bf 52.5} & {101.3} & {14.6} & 66.0 & 41.3 & {\bf 13.8} \\
    & Ours & {\bf 84.0} & {\bf 52.5} & {\bf 99.6} & {\bf 12.7} & {65.6} & {\bf 41.0} & {13.9} \\
    \bottomrule
    \end{tabular}}
    \end{threeparttable}
    \vspace{1pt}
    \caption{Evaluation on 3DPW and Human3.6M dataset with the SMPL annotations of Human3.6M. ``Ours\ddag" denotes that we directly take the orientation normalized motion as input and output without denoising training scheme.}
    \vspace{-1mm}
    \label{tab:mesh_construct}
\end{table*}

\begin{table}
  \small
  \centering
  \begin{threeparttable}
  \resizebox{\linewidth}{!}{
    \begin{tabular}{lcccc}
    \toprule
    Method& MPJPE $\downarrow$ & PA-MPJPE $\downarrow$ & MPVPE $\downarrow$ & Accel Error $\downarrow$ \cr
    \cmidrule(lr){1-5}
    VIBE~\cite{kocabas2020vibe} & 91.9 & 57.6 & - & 25.4 \\
    MEVA~\cite{luo20203d} & 86.9 & 54.7 & - & {11.6} \\
    TCMR~\cite{choi2021beyond} & 86.5 & {\bf 52.7} & 103.2 & {\bf 6.8} \\
    \cmidrule(lr){1-5}
    Ours\ddag & {85.5} & {53.6} & {102.6} & 15.9\\
    Ours & {\bf 85.2} & {53.2} & {\bf 102.1} & 14.3\\
    \bottomrule
    \end{tabular}}
    \end{threeparttable}
    \vspace{1pt}
    \caption{Evaluation on 3DPW without the SMPL annotations of Human3.6M. ``Ours\ddag" means without denoising training scheme.}
    \label{tab:mesh_construct_noh36m}
\end{table}

\subsection{Skeleton-based Action Recognition}
\textbf{Problem definition.} Compared with SMPL, skeleton is more general to application and easy to access. Thus, we integrate our learnt motion prior into the skeleton-based action recognition. This task aims to classify a motion sequence $x_i = \{x_i^t\}_{t=1}^N$, where $x_i^t\in\mathbb{R}^{J\times 3}$ represents the position of $J$ joints, into a certain category $y_i\in\mathcal{Y}$.

\textbf{Architecture.} First, we freeze the pre-trained decoder with parameters $\mathbf{\vartheta}$, which has specified the representation space with posterior density $p_\mathbf{\vartheta}(\mathbf{z}|\mathbf{x})$ over representation $\mathbf{z}$~\cite{kingma2013auto}. Then, to adapt to the skeleton-based motion, which spans a subspace of the SMPL-based data, and embed them into pre-defined representation space, we remove the SMPL parameters, $\hat{\theta^g}$ and $\theta^l$, in $\Phi$ (see Sec.~\ref{subsec:motion_prior_framework}) and retrain the encoder from scratch with only skeleton information as input. Note that, this process follows the VAE training paradigm and it is agnostic to the action recognition task and dataset.

With the retrained and frozen encoder, each skeleton-based motion $x_i$ is embedded into the representation $z_i$ in our pre-defined prior space for further recognition. Then, we feed the representation to a three-layer multi-layer perceptron (MLP) with ReLU activation and $1,024$ hidden units, and it outputs classification logits for final prediction.

\section{Experiments}
In Sec.~\ref{subsec:motion_prior_exp}, we introduce the dataset and implementation details in our motion prior, and evaluate the performance quantitatively and qualitatively in Sec.~\ref{subsec:motion_prior_eval}.
Then, in Sec.~\ref{subsec:mesh_rec_exp} and~\ref{subsec:motion_pred_exp}, we integrate our method into the \textit{human motion reconstruction} and \textit{motion prediction} task to show that we provide the efficient motion representation space for inverse problem with probability as cues.
Also, we exploit the \textit{skeleton-based action recognition} to show that the learnt representation is distinguishable in Sec.~\ref{subsec:ar}. To prove that our prior encodes the reasonable transition between poses, we conduct evaluation on \textit{motion infilling} task in Sec. \ref{subsec:motion_infilling_exp}.

\textbf{Metric.}  Following~\cite{kanazawa2019learning,kocabas2020vibe,luo20203d}, to evaluate the performance, we mainly use the following metrics: mean per joint position error (MPJPE), MPJPE after procrustes-alignment (PA-MPJPE) and mean per vertex position error (MPVPE), which are measured in $(mm)$. Besides, acceleration error in $(mm/s^2)$ is used to measure the smoothness. 

\subsection{Motion Prior Implementation}
\label{subsec:motion_prior_exp}
\textbf{Dataset.} Following~\cite{luo20203d}, we train our motion prior with a large and varied database of human motion that unifies different mocap datasets, AMASS~\cite{mahmood2019amass}, and split the original dataset into train/val set with pre-processing (details are in Sup. Mat.). To evaluate the generalization ability and effectiveness of our motion prior, we show the VAE reconstruction performance on the unseen in-the-wild {3DPW}~\cite{von2018recovering}.

\textbf{Implementation.} In the training stage, weights in Eq. (\ref{eq:total-loss}) are set as $\{\lambda_{rec}, \lambda_{kl}, \lambda_{vposer}\}=\{1, 0.01, 0.001\}$. We utilize the Adam optimizer with the learning rate $1e^{-4}$ and weight decay $1e^{-4}$. The network is trained for $250$ epochs with the batch size of $60$.
\label{subsec:visual_prior_space_exp}
\begin{figure}
    \centering
    \vspace{-3mm}
    \includegraphics[width=0.99\linewidth]{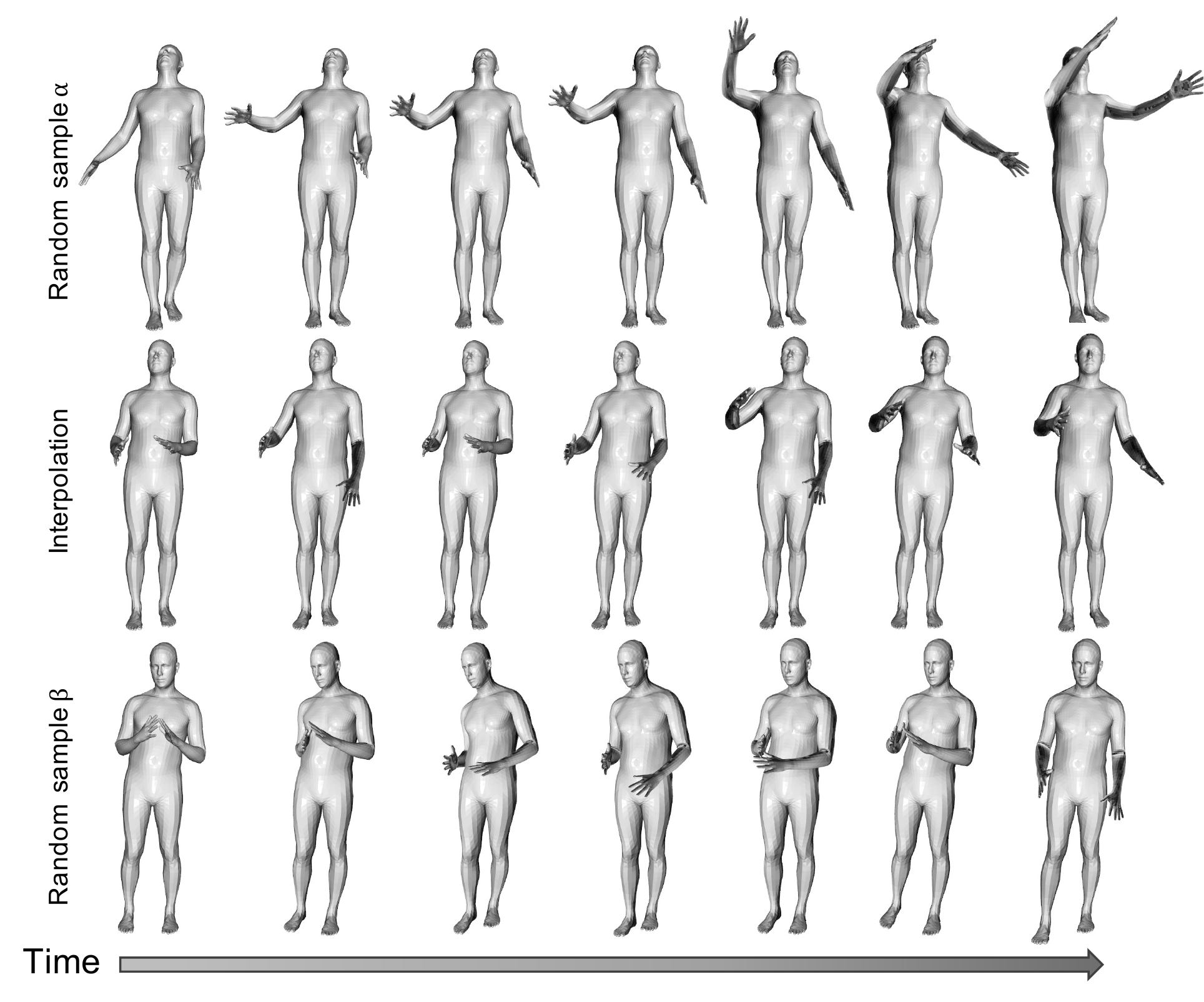}
    \caption{Illustration of sampled motion in top and bottom rows and interpolated motion in the middle row from our prior latent space. These consecutive poses selected from the first 60 frames of the generated motion with an interval of 10 frames.}
    \label{fig:visual_sample}
    \label{fig:my_label}
\end{figure}
\subsection{Motion Prior Evaluation}
\label{subsec:motion_prior_eval}
To demonstrate the effectiveness of our proposed methods, we conduct experiments on the test set of 3DPW. The VAE reconstruction error reported in Tab. \ref{tab:prior_ablation} shows that our prior generalizes well from AMASS to the unseen 3DPW, which is important for the versatility.
Then, we train a vanilla VAE with the global orientation normalization. As shown in Tab. \ref{tab:prior_ablation}, compared with MEVA~\cite{luo20203d} which reduces the complexity of data space by resorting to shorter-term motions, we achieve the better performance and it demonstrates the effectiveness of our global orientation normalization. Also, the proposed frequency guidance further improves the performance of vanilla VAE, and it proves that sequence-based and segment-based frequency guidance are effective. Compared with sequence-based frequency that indicates the category information mainly determined by the local poses, segment-based frequency focuses on both orientation transition and local poses compression, leading to better tradeoff between MPJPE and PA-MPJPE. 

To qualitatively show that we construct an expressive prior space for plausible motions, we randomly sample the latent variable $z_{mot}\in\mathbb{R}^{256}$ from the normal distribution and generate the human motion. The top and bottom rows in Fig. \ref{fig:visual_sample} show two motions generated from sampled latent variables $z_{mot}^\alpha$ and $z_{mot}^\beta$. Also, we average these two variables and get the interpolated motion in the middle row of Fig. \ref{fig:visual_sample}. These reasonable results demonstrate our prior is plausible while tractably and continuously distributed.

\subsection{Human Motion Reconstruction}
\label{subsec:mesh_rec_exp}
\textbf{Dataset.} Following~\cite{kocabas2020vibe}, in training phase, we use the {InstaVariety}~\cite{kanazawa2019learning} dataset to provide pseudo ground-truth 2D annotations. Also, we utilize {3DPW} and {Human3.6M}~\cite{ionescu2013human3} for SMPL parameters supervision, while employing {MPI-INF-3DHP}~\cite{mehta2017monocular} for 3D joints supervision. For evaluation, we show results on the test set of {3DPW} and {Human3.6M}. Specifically, on Human3.6M, we use [S1, S5, S6, S7, S8] as the training set and [S9, S11] as the test set.

\begin{figure*}
    \centering
    \includegraphics[width=0.99\linewidth]{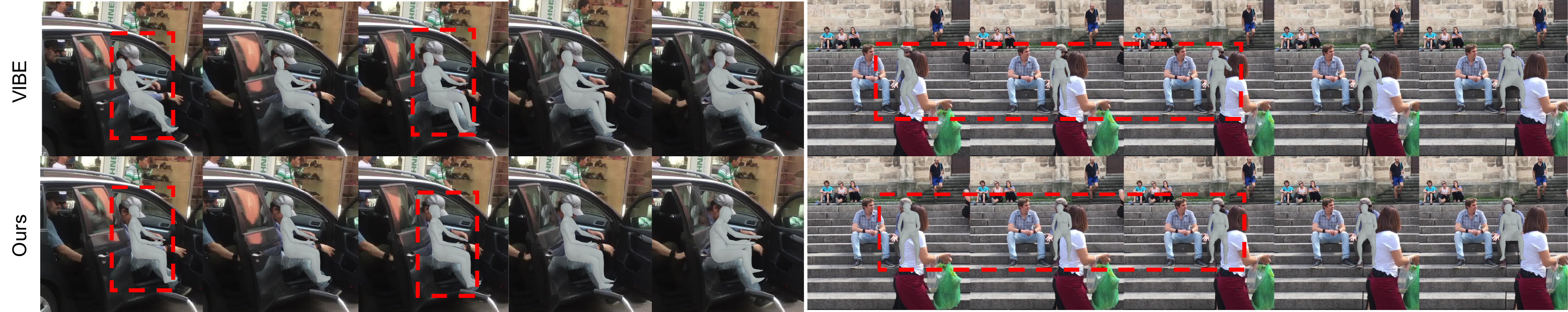}
    \caption{Qualitative comparison between VIBE (top) and our method (bottom) on the in-the-wild 3DPW.}
    \vspace{-1mm}
    \label{fig:3dpw_viz}
\end{figure*}

\textbf{Experimental results.} As introduced in Sec. \ref{subsec:3d-reconstruct}, we take the VIBE~\cite{kocabas2020vibe} as our backbone while keeping the same setting, \eg, the length of video $T=16$. 
Tab.~\ref{tab:mesh_construct} shows the quantitative results compared with the state-of-the-art methods on {3DPW} and {Human3.6M}. 
Compared with VIBE, our motion prior improves the smoothness and reduces the acceleration error from $27.0mm/s^2$ to $12.7m/s^2$ on {3DPW} and from $27.3mm/s^2$ to $13.9mm/s^2$ on {Human3.6M}. Also, because our motion prior naturally encodes the reasonable transition between poses and provides the context information, the reconstruction error (\eg, MPJPE, PA-MPJPE and MPVPE) is also improved. Fig. \ref{fig:3dpw_viz} illustrates the qualitative results in presence of occlusion. Compared with the adversarial prior in VIBE, which only offers a plausible prediction, our motion prior generates a predicted motion with higher probability and achieves better results. More qualitative results are provided in Sup. Mat.

Furthermore, following~\cite{luo20203d}, we also show the performance without SMPL parameters of Human3.6M and the results are shown in Tab.~\ref{tab:mesh_construct_noh36m}. Compared with previous works which are carefully designed for reconstruction task and output the prediction in a frame-wise manner, the MPJPE and MPVPE are improved. Specifically, compared with MEVA~\cite{luo20203d} that constructs the motion prior with more complex latent space and shorter-term motion, the improvement also demonstrates the efficiency of our motion prior.

\textbf{Effectiveness of denoising scheme.} Furthermore, we also conduct experiments to prove the effectiveness of our proposed denoising training scheme. From Tab.~\ref{tab:mesh_construct} and \ref{tab:mesh_construct_noh36m}, we can see that the performances are improved on both two settings, especially in MPJPE, which shows that the denoising scheme of rotation noise helps to learn a better representation of orientation transition between frames. 

\subsection{Motion Prediction}
\label{subsec:motion_pred_exp}
\textbf{Dataset.} Following~\cite{zhang2019predicting}, we train our network on the combination of {InstaVariey}, {PennAction}~\cite{zhang2013actemes} and Human3.6M. Specifically, the Human3.6M is split into train/val/test set as [S1, S6, S7, S8]/[S5]/[S9, S11].

\textbf{Experimental results.} In Sec.~\ref{subsec:motion_predict}, we introduce that PHD~\cite{zhang2019predicting} is taken as our backbone and we also use the past $T=15$ frames as input and train the network to predict future $25$ frames. 
Then, we discard the Dynamic Time Warping in~\cite{zhang2019predicting} and compare with them under the same setting. As shown in Tab.~\ref{tab:motion_pred}, the performance of first $20$ frames are improved with our prior. Following~\cite{zhang2019predicting}, we also report result of the $30th$ frame, which is not supervised in the training phase and directly taken from the motion generated by $z_{mot}$, and we can see that our motion prior still improve the result, because it naturally encodes a sequence of plausible motion starting from the given frames.

\begin{table}
  \small
  \centering
  \begin{threeparttable}
    \resizebox{\linewidth}{!}{
    \begin{tabular}{lcccccc}
    \toprule
    \multirow{2}{*}{Method}&
    \multicolumn{5}{c}{PA-MPJPE $\downarrow$}\cr
    \cmidrule(lr){2-6}
    & 1$th$~ & 5$th$~ & 10$th$~ & 20$th$~ & 30$th$~\cr
    \midrule
    Zhang~\etal~\cite{zhang2019predicting}~~ & 57.7~ & 61.2~ & 64.4~ & 67.1~ & 81.1~\\
    \midrule
    Ours & {\bf 51.9~} & {\bf 61.1~} & {\bf 63.3~} & {\bf 63.9~} & {\bf 80.2~} \\
    \bottomrule
    \end{tabular}}
    \end{threeparttable}
    \vspace{1pt}
    \caption{Results of motion prediction from video without Dynamic Time Warping. We report the PA-MPJPE for the 1$th$, 5$th$, 10$th$, 20$th$, 30$th$ frame in the future motion.}
    \label{tab:motion_pred}
\end{table}

\subsection{Action Recognition}
\label{subsec:ar}
\begin{table}[]
    \centering
    \resizebox{\linewidth}{!}{
    \begin{tabular}{clcccc}
        \toprule
        \multirow{2}{*}{Loss}&\multirow{2}{*}{Method}&
    \multicolumn{2}{c}{BABEL-60}&\multicolumn{2}{c}{BABEL-120}\cr
    \cmidrule(lr){3-4} \cmidrule(lr){5-6}
    & & Top1$\uparrow$ & \hspace{-2mm}Top1-\textit{norm}$\uparrow$ & Top1$\uparrow$ & \hspace{-2mm}Top1-\textit{norm}$\uparrow$\cr
        \midrule
        \multirow{3}{*}{\rotatebox{90}{\begin{tabular}{c}CE\end{tabular}}} & 2s-AGCN~\cite{BABEL_CVPR_2021} & {\bf 44.9} & {17.2} & {\bf 43.6} & 11.3 \\
        & Ours\dag & 38.6 & 22.4 & 36.0 & 17.4\\
        & Ours & 40.3 & {\bf 23.6} & 37.8 & {\bf 18.2} \\
        \midrule
        \multirow{3}{*}{\rotatebox{90}{\begin{tabular}{c}Focal\end{tabular}}} & 2s-AGCN~\cite{BABEL_CVPR_2021} & 37.6 & 25.7 & {\bf 31.7} & 19.2 \\
        & Ours\dag & 32.7 & 26.2 & 30.4 & 22.3 \\
        & Ours & {\bf 38.1} & {\bf 27.2} & 31.5 & \textbf{25.5} \\
        \bottomrule
    \end{tabular}}
    \vspace{1pt}
    \caption{Evaluation on BABEL dataset. ``Ours" means that we only train the MLP and freeze the pre-trained encoder. ``Ours\dag" denotes that we train the whole framework from scratch.}
    \label{tab:action_recognition}
\end{table}
\textbf{Dataset.} As introduced in Sec.~\ref{sec:related_work}, BABEL~\cite{BABEL_CVPR_2021} provides more diversity and long-tailed distribution of samples, that is more close to real-world applications. Therefore, we conduct experiments on BABEL dataset and follow the official split in~\cite{BABEL_CVPR_2021} to use the long-tailed BABEL-60 and BABEL-120, containing $60$ and $120$ action categories, respectively.

\textbf{Experimental results.} 
In Tab.~\ref{tab:action_recognition}, we report two metrics: Top1 and Top1-\textit{norm} accuracy (the mean Top1 across categories). Compared with Top1, Top1-\textit{norm} better reveals the performance of tackling the long-tailed distribution problem. In addition, following~\cite{BABEL_CVPR_2021}, we use both cross-entropy and focal loss in the training phase. On BABEL-60 and BABEL-120, our method achieves better performance in Top1-\textit{norm} compared with the baseline, which is end-to-end trained on the dataset. It demonstrates the effectiveness and generalization ability of learnt representation.
Also, we retrain the skeleton encoder together with MLP in Sec.~\ref{subsec:ar} from scratch in an end-to-end way. 
As shown in Tab.~\ref{tab:action_recognition}, the result is worse, and it is because, decoupling representation learning may help to retain more distinguishable information and lead to more generalizable representation~\cite{kang2019decoupling}. 

\subsection{Motion Infilling}
\label{subsec:motion_infilling_exp}
\begin{table}
\small
    \centering
    \begin{tabular}{lcc}
        \toprule
        Method & 60 Frames $\downarrow$ & 120 Frames $\downarrow$ \\
        \midrule
        Interpolation & 10.45 ($\pm$ 15.5) ~~ & ~~ 17.04 ($\pm$ 24.4) \\
        Holden~\etal~\cite{holden2016deep} & 15.28 ($\pm$ 19.1) ~~ & ~~ 18.26 ($\pm$ 24.5)\\
        Kaufmann~\etal~\cite{kaufmann2020convolutional} & 4.96 ($\pm$ 8.5) ~~ & ~~ 12.00 ($\pm$ 19.5)\\
        \midrule
        Ours &  {\bf 2.01 ($\pm$ 2.15)} ~~ & ~~ {\bf 2.51 ($\pm$ 2.52)}\\
        \bottomrule
     \end{tabular}
    \vspace{2mm}
    \caption{Results of motion infilling tasks. 3D joint errors are reported by the mean and standard deviation in \textit{cm} computed over all joints and frames on the validation set.}
    \label{tab:motion_infill}
\end{table}

To show that our prior encodes the transition between poses, we exploit the motion filling task, which aims to fill in missing frames in a human motion. Instead of designing a network, we utilize our pre-trained decoder with fixed weights and perform the motion infilling in an optimizing manner. We refer the reader to Sup. Mat. for more details.

\textbf{Dataset.} We use the dataset released by~\cite{holden2016deep}, where each pose is represented as $22$ joints. Following~\cite{kaufmann2020convolutional}, we evaluate the performance on the validation set, and $T$ frames are randomly selected as the missing frames in each motion.

\textbf{Experimental results.} Because the data is represented in skeleton, we regress the $22$ joints from predicted SMPL model so as to optimize the $z_{mot}$. Tab.~\ref{tab:motion_infill} shows the results in two settings: i) $T=60$ and ii) $T=120$. We outperform previous methods trained on this dataset, which shows the generalization ability of our motion prior and proves that it naturally represents the inherent transition between poses. Also, Fig.~\ref{fig:motion_infilling} illustrates the qualitative results and the comparison between ground truth is in Sup. Mat.

\begin{figure}
    \centering
    \includegraphics[width=0.99\linewidth]{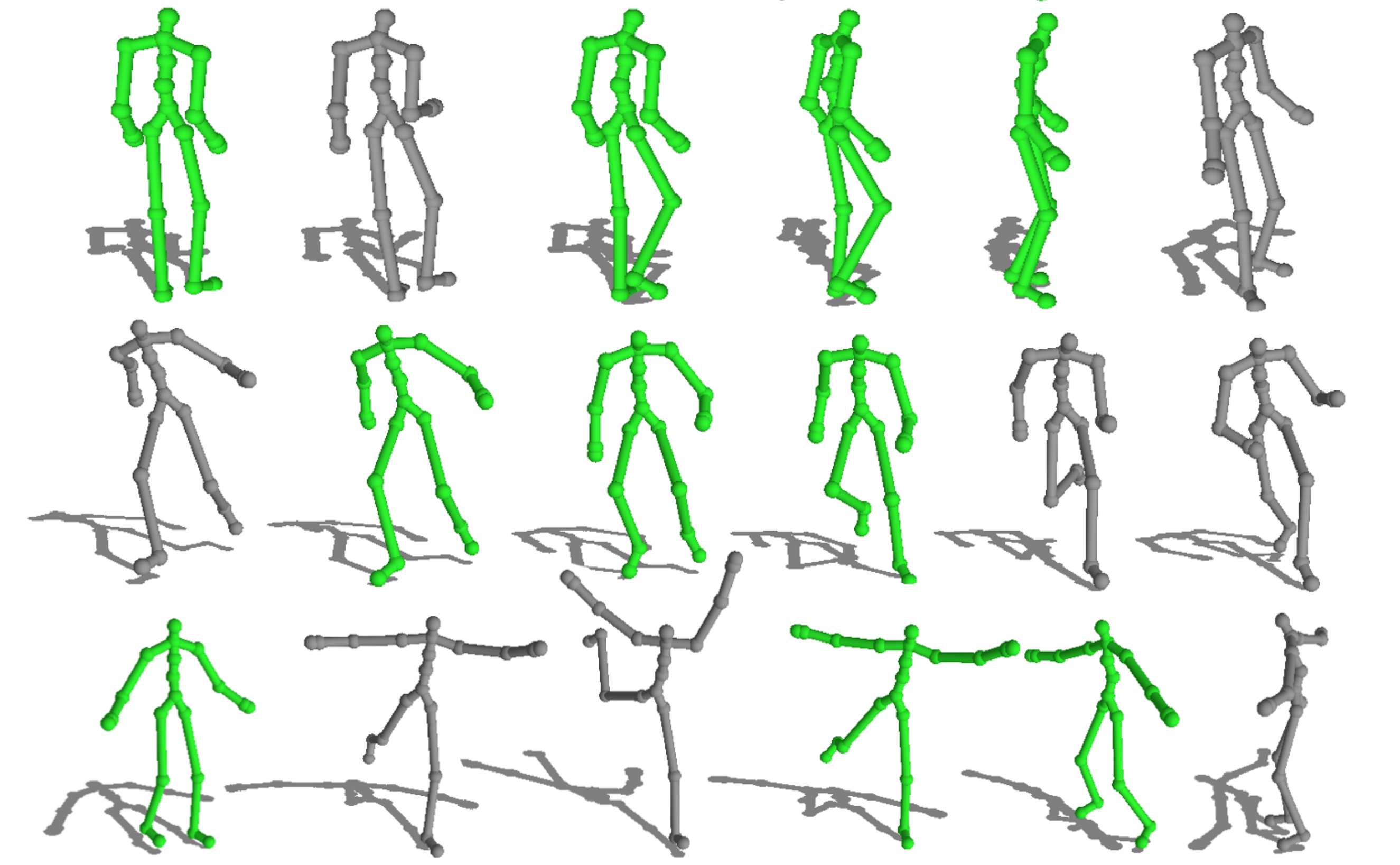}
    \caption{Illustration of motion infilling. We visualized six consecutive poses, with an interval of ten frames. Poses in gray mean the known frame and green ones denote the generated poses.}
    \label{fig:motion_infilling}
\end{figure}

\section{Conclusion}
In this paper, we summarize the indispensable properties of a motion prior and propose a versatile motion prior which models the inherent probability distribution of motions. To keep the learnt representation space efficient, we introduce a global orientation normalization and a two-level frequency guidance. Then, we adopt a denoising training scheme to provide the consistent and distinguishable representation for each motion. Finally, we embed our proposed motion prior into different prevailing backbones and conduct extensive experiments on different tasks. The results show that the motion prior can improve the baseline and achieve the state-of-the-art performance, and it demonstrates the versatility and effectiveness of our prior.

\noindent\textbf{Acknowledgments.}
This work is sponsored by the National Key Research and Development Program of China (2019YFC1521104), National Natural Science Foundation of China (61972157), Shanghai Municipal Science and Technology Major Project (2021SHZDZX0102), Shanghai Science and Technology Commission (21511101200), Art major project of National Social Science Fund (I8ZD22), National Natural Science Foundation of China under Grant (62176092), Shanghai Science and Technology Commission (21511100700) and SenseTime Collaborative Research Grant.

{\small
\bibliographystyle{ieee_fullname}
\bibliography{3dv}
}

\newpage
\ 

\newpage
\section*{Appendix}
\section{Data Preparation}
\label{sec:data_prep}
Because AMASS~\cite{mahmood2019amass} consists of diverse motion capture data with SMPL~\cite{loper2015smpl} parameters. Therefore, to provide a unified training and validation set, we first downsample the original sequence into $30$ fps and divide each sequence into sub-sequence with $K=128$ frames. The overlap between every two sub-sequences is $30$ frames. Second, we randomly select $15\%$ of the sub-sequence dataset as the validation set to evaluate the performance of our proposed motion prior, and ensure that there are no overlapping sub-sequences with the training set.

\begin{table}[!h]
    \small
    \centering
    \begin{tabular}{l|c}
    \toprule
        ~~Method & Parameters (M) \\
        \midrule
        ~~VIBE~\cite{kocabas2020vibe}~~ & 21.31\\
        ~~MEVA~\cite{luo20203d}~~ & 58.74\\
        ~~TCMR~\cite{choi2021beyond}~~ & 81.91\\
        \midrule
        ~~Ours & 28.14\\
        \bottomrule
    \end{tabular}
    \vspace{1mm}
    \caption{The amount of network parameters compared with the state-of-the-art methods.}
    \label{tab:model_complexity}
\end{table}

\section{Model Complexity}
\label{sec:model_complexity}
Here, we compare the model complexity of our method and previous works in Tab. \ref{tab:model_complexity}. As we can see, we achieve the better performance as well as maintain the low complexity of network.

\begin{table}[]
    \small
    \centering
    \begin{threeparttable}
    \resizebox{\linewidth}{!}{
    \begin{tabular}{l|cccc}
        \toprule
        Dim & MPJPE $\downarrow$ & PA-MPJPE $\downarrow$ & MPVPE $\downarrow$ & Accel Error $\downarrow$ \cr
        \midrule
        128 & 35.66 & 24.90 & 43.87 & 5.18\\
        256 & 26.01 & 17.44 & 32.70 & 5.07 \\
        512 & 27.25 & 16.79 & 33.42 & 5.01\\
        \bottomrule
    \end{tabular}}
    \end{threeparttable}
    \vspace{1mm}
    \caption{VAE reconstruction error on in-the-wild 3DPW dataset. ``Accel Error" stands for the acceleration error and the symbol ``$\downarrow$" denotes that the lower the result, the better.}
    \label{tab:prior_ablation}
\end{table}

\begin{figure}[!h]
    \centering
    \includegraphics[width=\linewidth]{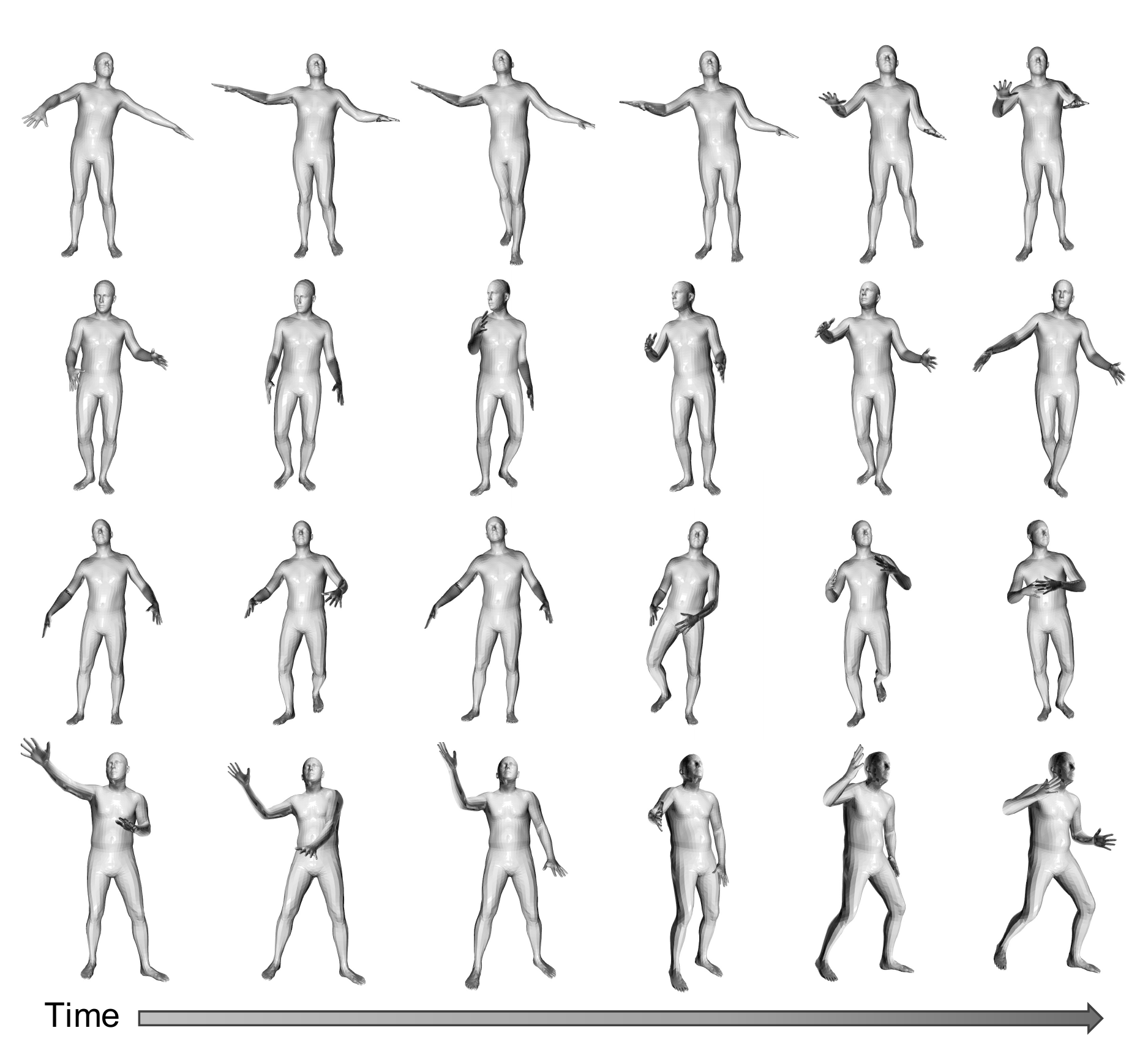}
    \caption{Illustration of sampled motion. These consecutive poses selected from the first 60 frames of the generated motion with an interval of 10 frames}
    \label{fig:sample_visualization}
\end{figure}

\begin{figure*}
    \centering
    \includegraphics[width=0.98\linewidth]{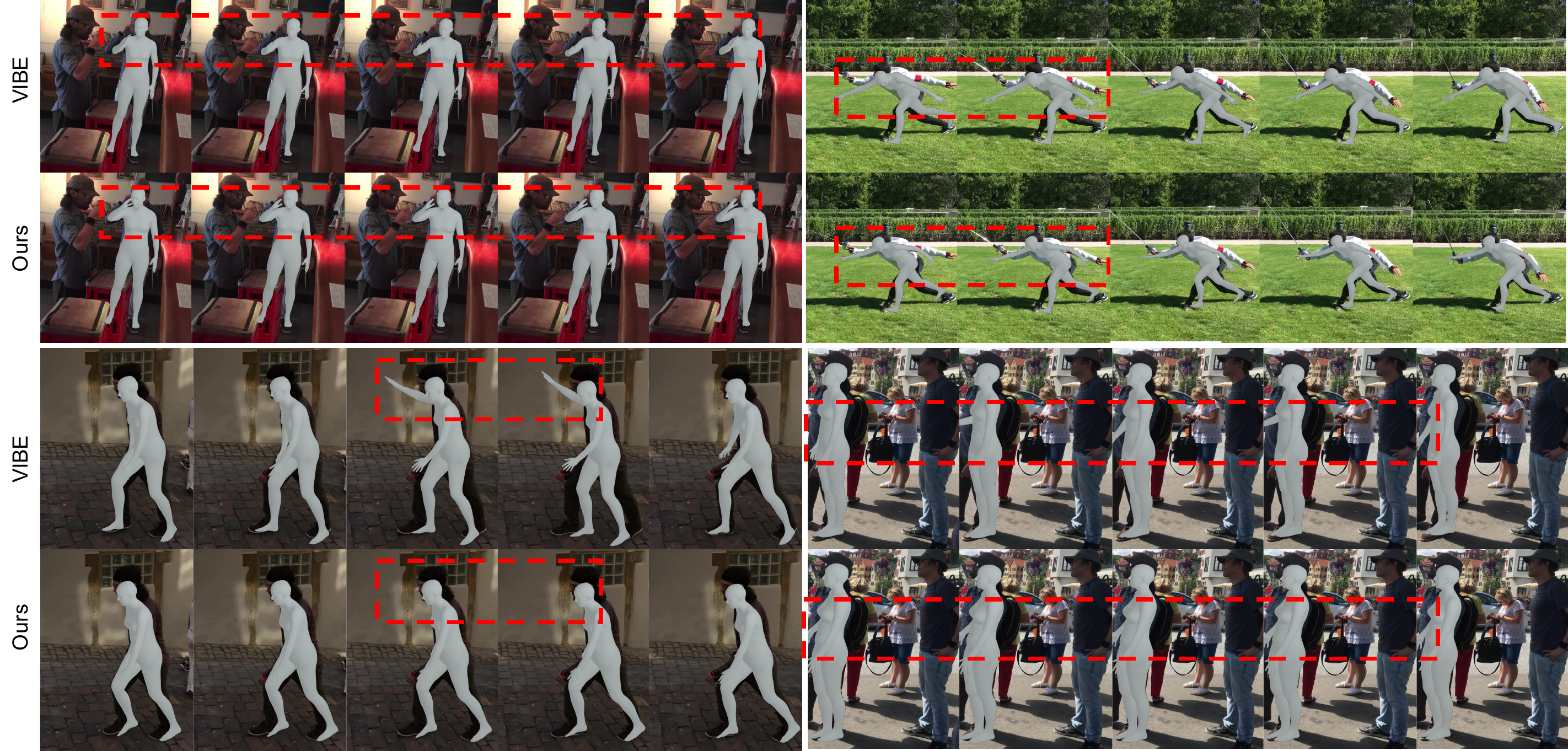}
    \caption{Qualitative comparison between VIBE and our method on the in-the-wild 3DPW.}
    \label{fig:motion_reconstruction}
\end{figure*}

\begin{figure*}
    \centering
    \includegraphics[width=0.98\linewidth]{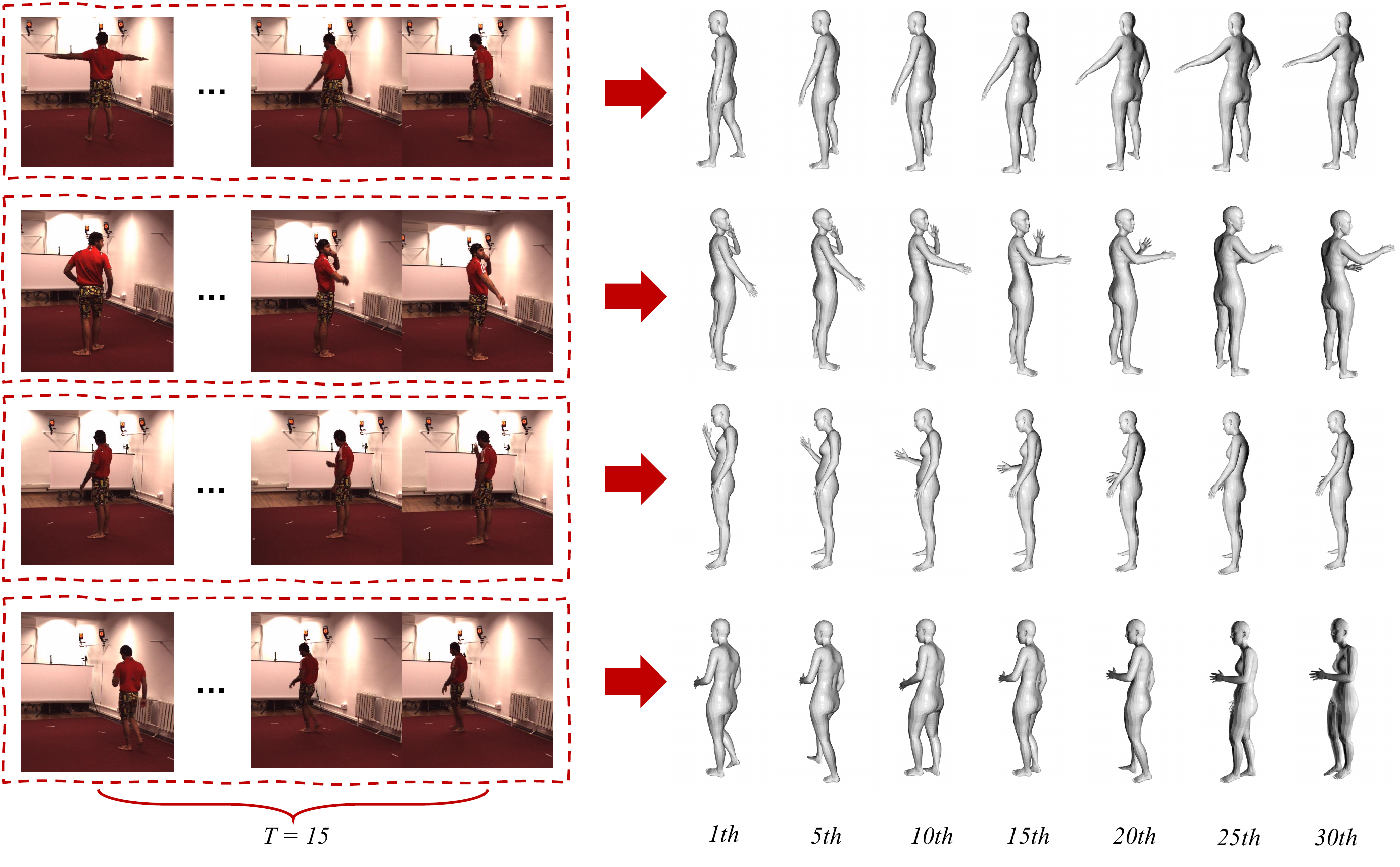}
    \caption{Visualization of motion prediction results on the Human3.6M dataset. Results are shown at every 5 frames.}
    \label{fig:motion_prediction_fig}
\end{figure*}

\begin{figure*}
    \centering
    \includegraphics[width=\linewidth]{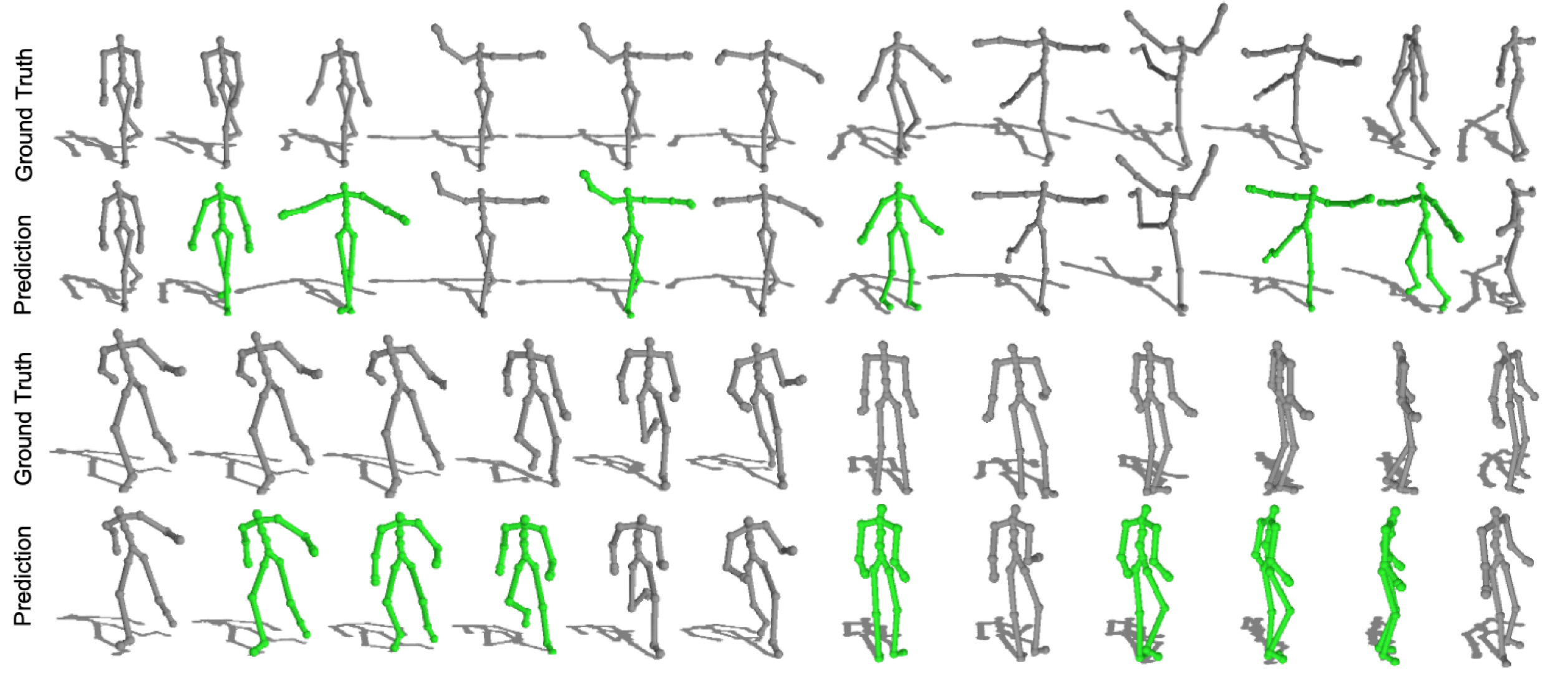}
    \caption{Illustration of motion infilling. We visualized six consecutive poses, with an interval of ten frames. Poses in gray mean the known frame and green ones denote the generated pose.}
    \label{fig:motion_infilling}
\end{figure*}

\section{Number of Representation Dimension}
\label{sec:num_dim}
To show the impact of the number of representation dimension, we conduct the ablation study and results are shown in Tab. \ref{tab:prior_ablation}. As we can see, compared with the $256$-dimensional space that we adopt in our motion prior, the $128$-dimensional latent space has the inferior performance. Because the expressive ability is limited. Furthermore, the $512$-dimensional representation space does not significantly improve the performance. Because it is hard for the high-dimensional latent space to balance between maintaining favorable performance and following the normal distribution regularization. Therefore, we choose to construct our motion prior with the $256$-dimensional latent space which provides both satisfied performance and compact representation space.

\section{Implementation in Motion Infilling}
\label{sec:motion-infilling-imple}
In the motion infilling task, a 3D human motion sequence $\mathbf{X}^{'}$ with missing frames is given. Using our pre-trained decoder with fixed weights, we aim to complete the missing frames by directly optimizing a zero-initialized latent variable $z_{mot}$ to generate the human motion. Then, we use the available frames in $\mathbf{X}^{'}$ to supervise the corresponding frames in the output motion and optimize for 30 iterations with learning rate $0.2$.

\section{Qualitative Results}
\label{sec:qualitative_results}
Fig.~\ref{fig:sample_visualization} visualizes the motion randomly sampled from our motion prior. Then, in Fig. \ref{fig:motion_reconstruction}, we compare the visualization results with VIBE~\cite{kocabas2020vibe} on 3DPW~\cite{von2018recovering} dataset, so as to show that embedding our motion prior to VIBE can improve the performance. We suggest that you can zoom in for more details. Additionally, we upload some video demo \href{https://youtu.be/MLmCq0f-RP8}{here}. For fair comparison, we use the same roughly estimated camera for 2D rendering of both VIBE and our results. In Fig. \ref{fig:motion_prediction_fig}, we visualize the predicted future motion results given $T=15$ past frames on the Human3.6M dataset~\cite{ionescu2013human3}. Fig. \ref{fig:motion_infilling} illustrates the comparison between ground truth and predicted results of motion infilling. Note that, even given the known frames as the supervision, the predicted poses in the known frames may still different from the ground truth.

\end{document}